\documentclass[pmlr]{jmlr}

\usepackage{multirow}
\usepackage{longtable}
\usepackage{booktabs}
\usepackage[load-configurations=version-1]{siunitx} 
\usepackage{graphicx}
\usepackage{epstopdf}

\theorembodyfont{\upshape}
\theoremheaderfont{\scshape}
\theorempostheader{:}
\theoremsep{\newline}

\jmlrvolume{1}
\jmlryear{2023}
\jmlrworkshop{perprint}

\title[Concept Prerequisite Relation Prediction]{Concept Prerequisite Relation Prediction by Using \titlebreak Permutation-Equivariant Directed Graph Neural Networks
\titletag{\thanks{
This work was supported in part by National Natural Science Foundation of China (62272392, U22A2025) and the funding for Teaching \& Learning Reform at NPU (2023JGZ14).}}
}

  \author{\Name{Xiran Qu} \Email{xrqu@mail.nwpu.edu.cn}\\ 
   \Name{Xuequn Shang} \Email{shang@nwpu.edu.cn}\\
   \Name{Yupei Zhang\nametag{\thanks{Corresponding Author}}} \Email{ypzhaang@nwpu.edu.cn}\\
   \addr School of Computer Science, Northwestern Polytechnical University, Xi'an, 710129, China.\\
  \addr Big data Storage and Management MIIT Lab, Xi'an, 710129, China.}

\editor{Editor's name}

\begin{document}

\maketitle

\begin{abstract}
This paper studies the problem of CPRP, concept prerequisite relation prediction, which is a fundamental task in using AI for education. CPRP is usually formulated into a link-prediction task on a relationship graph of concepts and solved by training the graph neural network (GNN) model. However, current directed GNNs fail to manage graph isomorphism which refers to the invariance of non-isomorphic graphs, reducing the expressivity of resulting representations. We present a permutation-equivariant directed GNN model by introducing the Weisfeiler-Leman test into directed GNN learning. Our method is then used for CPRP and evaluated on three public datasets. The experimental results show that our model delivers better prediction performance than the state-of-the-art methods.
\end{abstract}
\begin{keywords}
Concept Prerequisite Relation, permutation-equivariant GNNs, Weisfeiler-Leman Test, Directed Graph Learning, AI for Education.
\end{keywords}

\section{Introduction}
With the continuous advancement of dissemination methods, an increasing number of educational resources are becoming available for people to learn \cite{fischer2020mining}. Therefore, finding prerequisite relationships among concepts has become an important issue requiring investigation in the field of AI for education \cite{pan2017prerequisite,roy2019inferring}. Generally, this Concept Prerequisite Relation Prediction task, CPRP, is modeled as the link prediction problem in many studies \cite{sun2022conlearn,roy2019inferring}. Applications of CRPR involve material recommendation \cite{guan2023kg4ex}, learning path planning \cite{shi2020learning}, and optimization of problem-solving paths \cite{le2023knowledge}.

There are numerous approaches to solving the link prediction problem, including probabilistic models, spectral clustering, evolutionary algorithms, and deep graph learning models \cite{kumar2020link}. Currently, graph neural networks (GNNs) have become the benchmark method and presented state-of-the-art performance in link prediction \cite{cai2021line}, including many GNN-based solutions to CPRP \cite{roy2019inferring,jia2021heterogeneous,sun2022conlearn,mazumder2023graph}.
To improve the capability of GNNs, 
\cite{long2022pre} and \cite{chamberlain2022graph} proposed to pre-training node features through the method of graph reconstruction and utilizing subgraph sketches to pass messages in subgraph GNNs, aiming to enhance the capabilities of GNNs and reduce computational costs.
Thess models mitigated the expressive limitations, such as the inability to count triangles and distinguish automorphic nodes. Persistent homology was also adopted in the work of \cite{yan2021link} to extract topological information from graphs, which was integrated with node features to enhance the expressive power of GNNs for link prediction. Besides, the Weisfeiler-Leman test was introduced by \cite{morris2019weisfeiler} to improve the capability of differentiating graph isomorphism in undirected GNN learning \cite{huang2022going}.

However, the CPRP problem is usually formulated into the directed-link prediction in a directed graph \cite{sun2022conlearn}. Rather than the undirected graph, the edges can indicate the prerequisite relation, describing the flow of information from one node to another. The difference makes undirected GNNs not directly applicable to directed graphs. Hence, \cite{salha2019gravity} designed a new gravity-inspired decoder to extend the graph Autoencoders, while \cite{wu2019session} adopted two weight matrixes for forward edges and backward edges, respectively, to perform message passing between graph nodes. Nevertheless, there is a lack of studies on improving the expressive power of direct GNNs for CPRP.

In this paper, we extended the Weisfeiler-Leman test-based GNN model, i.e., SpeqNets recently introduced by \cite{morris2022speqnets}, into the directed graphs for CPRP. The proposed framework contributes to learning the permutation-equivariant direct GNNs and improving the prediction performance of CPRP by distinguishing non-isomorphism graphs. Experimental results on three public datasets manifest that our method performs better than the state-of-the-art methods \cite{sun2022conlearn,jia2021heterogeneous,roy2019inferring}.

\section{CPRP: Concept Prerequisite Relation Prediction}
The problem of CPRP refers to predicting prerequisite relations between knowledge concepts involved in learning \cite{sun2022conlearn}. For example, one should learn the knowledge concept (KC) of “conditional probability distribution" before learning “Bayesian theory." As usual, CPRP can be formulated into the directed-link prediction in a KC graph. 

\subsection{Problem formulation}
Denote by $G=(V,E)$ a directed KC graph with a vertex set $V = \{v_1,v_2,\ldots,v_N\}$, where $v_i$ represents a KC in our study, and an edge set $E=\{e_{ij}\}_{1\leq i,j \leq N}$, where $e_{ij}$ is the prerequisite relation $v_i \to v_j$ meaning that the KC $v_i$ is a prerequisite KC $v_j$. Denote by $A\in R^{N\times N}$ the adjacency matrix of $G$ with each element being 0 or 1. CPRP on $G$ can be written into
\begin{equation}\label{myeq1}
    \mathcal{P}(v_p,v_q)=\mathcal{M}(\mathcal{G}(v_p),\mathcal{G}(v_q))
\end{equation}
where $v_p$ and $v_q$ are two KCs; $\mathcal{P}$ is the probability of whether the relation $e_{pq}$ exists; $\mathcal{G}$ is a representation model that integrates KC information into a vector; the function $\mathcal{M}$ aims to obtain the existing probability of the prerequisite relation $e_{pq}$.

\subsection{Previous Methods}
Early works of CPRP usually extracted handcrafted features for $\mathcal{G}$, such as the contextual and structural features \cite{pan2017prerequisite}, while recent works are focused on designing deep-learning models, such as Siamese networks \cite{roy2019inferring} and GNNs \cite{sun2022conlearn}. 

Inspired by the promising performance, we implemented $\mathcal{G}$ by training a GNN model in this study. GNNs aim to learn node representations in a graph by iteratively aggregating the neighborhood features. In each layer, the feature of the node $v$ is updated by merging the information transmitted from its neighbors, expressed as
\begin{equation}
    \mathbf{f}^{(t)}_v=\mathcal{F}_{mer}^{W_1^{(t)}}(\mathbf{f}^{(t-1)}_v,\mathcal{F}_{agg}^{W_2^{(t)}}(\{{\mathbf{f}^{(t-1)}(u)}|u\in \mathcal{N}(v)\}))
\end{equation}
where $ \mathbf{f}^{(t)}_v$ indicates the feature vector of the node $v$ at the $t$-th layer; $\mathcal{F}_{mer}^{W_1^t}$ is the merging function with learned parameters $W_1^{(t)}$ and $\mathcal{F}_{agg}^{W_2^{(t)}}$ is the aggregating function with network parameters $W_2^{(t)}$; $\mathcal{N}(v)$ delivers the neighbors of node $v$ \cite{wu2020comprehensive}.

With the node representations from GNNs, many methods can estimate the prerequisite relations by employing similarity metrics or classical classifiers for $\mathcal{M}$ \cite{liang2018investigating}. However, the previous studies fail to consider the problem of graph isomorphism of the KC graphs, leading to low expressive powers for CPRP.



\section{Our CPRP Method}
\begin{figure}
\centering
\includegraphics[scale=0.56]{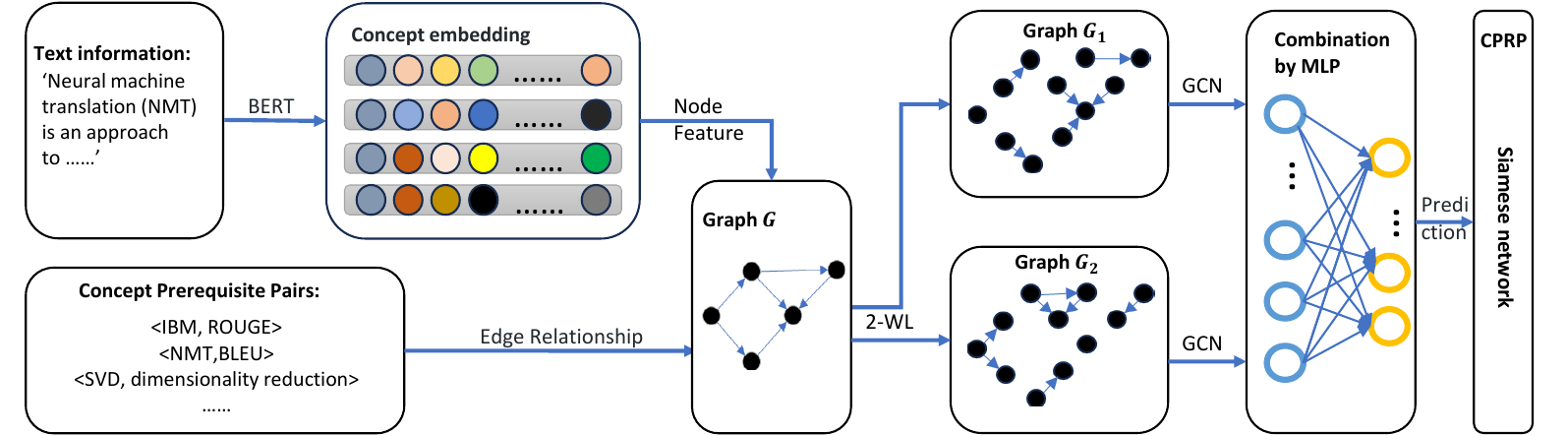}
\caption{Workflow of the proposed method for CPRP.}
\label{myfig1}
\end{figure}

To achieve fine representations of KCs, our method adopts the well-known Weisfeiler-Leman test \cite{morris2021power} to guide the GNN training in the KC graph $G$. With the obtained KC representations, the Siamese network computes the link probability, shown in Fig. \ref{myfig1}.

\subsection{Weisfeiler-Leman Test}
Denote by $S=\{(s_1,s_2,...,s_k)|s_i \in V,i\in I_k\}$ a $k$-tuple $S$ of vertices in the KC graph $G$, where $I_k$ indicates the first $k$ natural numbers; $s_i$ is the $i$-th element of $S$ specified to a $v_j \in V$. Let $V^k(G)$ be the collection of all $k$-tuples from the graph $G$. Weisfeiler-Leman (WL) test is to assign labels to each tuple in $V^k(G)$ and then iteratively relabel these tuples by merging their neighborhood labels. Here, the $j$-th neighborhood of the tuple $S$ is yielded by replacing its $j$-th element with every node $v \in V$, i.e., $\mathcal{N}_j^k(S,v) = \{(s_1,s_2,...,s_{j-1},v,s_{j+1},...s_k)|v \in V \}$ \cite{morris2021power}. Based on all neighborhoods of $S$, the WL test usually uses a predefined route to compute a new label for the merged node in the $i$-th iteration, i.e.,
\begin{equation} \label{myeq3}
    l_{i}(S)=\{\{(C^k_i(\mathcal{N}_1^k(S,v),\mathcal{N}_2^k(S,v),...,\mathcal{N}_k^k(S,v)|v\in V\}\}
\end{equation}
where $l_{i}(S)$ represents the obtained label and $C^k_i$ indicates a predefined function that maps all tuples $V^k$ to new labels. Then, the iterative labeling for $S$ can be expressed as 
\begin{equation}
    C^k_{i}(S) = Relabel((C^k_{i-1}(S),l_{i}(S)))
\end{equation}
where $Relabel(a,b)$ is to achieve $a \gets b$. 

The $k$-WL test is a powerful tool for distinguishing isomorphic graphs. Let $G$ and $H$ be two graphs. If the number of $k$-tuples with a specific label differs between graphs $G$ and $H$ at any iteration, then the two graphs are non-isomorphic. Specifically, 1-WL employs the 1-hop neighbors for $\mathcal{N}$ in Eq. (\ref{myeq3}). With the increase of $k$, the $k$-WL algorithm can become more capable of distinguishing non-isomorphic graphs.

\subsection{Main Steps}

\subsubsection{KC Graph Construction}
The first step of the proposed method is to achieve the node features for the given directed KC graph $G$ using the pre-trained BERT \footnote{https://huggingface.co/bert-large-uncased}. More specifically, the textual descriptions of KCs from the datasets or Wikipedia were obtained and fed into BERT to extract the KC embedding for $V$. With the given $E$, we achieved the KC graph $G$ \cite{devlin2018bert}. 

\subsubsection{Weisfeiler Leman Guided Directed GNNs}



On the resulting graph $G$, we proposed the directed GNN model guided by the $k$-WL test to achieve KC representations in this step. To be different from undirected graphs, we denote by $\mathcal{N}_{out}(v)$ the set of all out-neighbors of the node $v$. For two connected nodes $v$ and $w$ in a directed graph, if there exists an edge pointing from $v$ to $w$, node $w$ is defined as an out-neighbor of node $v$. In the proposed method, we redefined the $k$-tuple as follows,
\begin{equation}
    \widehat{V}(G)^k := \{(s_1,s_2,...,s_k) | s_i \in V, i\in I_k; \exists{j}\in I_{i-1}: {s_i}\in \mathcal{N}_{out}(s_j)\cup {\{s_j\}}\}. 
\end{equation}


The feature representation of $k$-tuple $(s_1,s_2,...,s_k)$ is here given by $[\mathbf{n_1}:\mathbf{n_2}:...:\mathbf{n_k}]$, where
$\mathbf{n_i}$ denotes the embedding of node $s_i$; 
$[\cdot:\cdot]$ denotes the concatenation of vectors. 
For $\forall {S}\in \widehat{V}(G)^k$, the $j$-th out-neighbor of ${S}$ can be cast as:
\begin{equation}
   \widehat{\mathcal{N}}^k_{j}({S}):=\{(s_1,s_2,...,s_{j-1},v,s_{j+1},...s_k)|v\in \mathcal{N}_{out}(s_j) \}
\end{equation}
For all $k$-tuples, the neighbor relationships generated on the $j$-th element are used to construct a graph representation as $G_j$. For all graphs $G_i(i\in I_k)$, each node in the graph $G_i$ represents a $k$-tuple in $\widehat{V}(G)^k$. For all graphs $G_i$, graph neural networks are constructed separately for training. After each layer of training is completed, the features of the same $k$-tuple on different graphs $G_i$ are fused using multilayer perceptron (MLP). The expression of $k$-tuple $S$ at the $t$-th layer is as follows:
\begin{equation}
    \widehat{\mathbf{f}}_S=\mathcal{F}_{MLP}(\mathcal{F}_S^{(1)},\mathcal{F}_S^{(2)},...,\mathcal{F}_S^{(k)})
\end{equation}
where $\widehat{\mathbf{f}}_S$ is defined as the representation of $k$-tuple $S$ at layer t. For $\forall i \in I_k$, $\mathcal{F}_S^{(i)}$ denotes the representation of $k$-tuple $S$ in graph $G_i$ after passing through the GCN \cite{kipf2016semi} designed for graph $G_i$. $\mathcal{F}_{MLP}$ denotes the multilayer perceptron. After several layers of learning, we obtain the representations of all $k$-tuples.


The representations of $k$-tuples obtained through GNN training are finally distributed to the nodes of $G$ using an average allocation scheme, shown as follows:
\begin{equation}
    \textbf{x}_j^{(i)}=\frac{1}{\# \{ \mathit{S}|\forall \mathit{S}\in \widehat{V}(G)^k,s_i=v_j\}}\displaystyle\sum_{\{ \mathit{S}|\forall \mathit{S}\in \widehat{V}(G)^k,s_i=v_j\}}^{}{\textbf{h}_S}
\end{equation}
where $s_i$ represents the $i$-th element in the $k$-tuple $S$. $\textbf{x}^{(i)}_j$ represents the transmission of feature expressions of $k$-tuples back to the representations of concept $v_j$, based on the $i$-th element in each $k$-tuple. $\textbf{h}_S$ denotes the feature of $k$-tuple $S$. $\# \{...\}$ denotes the number of elements in the set $\{...\}$. For $\forall i$, all $\textbf{x}^{(i)}_j$ vectors are merged and fed into a multi-layer perceptron to obtain node representations in $G$.
\begin{equation}
    \mathbf{f}_j = \mathcal{F}_{MLP}([\textbf{x}^{(1)}_j:\textbf{x}^{(2)}_j:...:\textbf{x}^{(k)}_j])
\end{equation}
where $\mathbf{f}_j$ represents the learned concept $v_j$ representations.
\subsubsection{Prediction Network}
After obtaining the KC representations, the Siamese network is employed here to predict the probability that the concept $v_p$ is a prerequisite concept of concept $v_q$ in Eq. (\ref{myeq1}). Let $\mathbf{\tilde e}_p$ and $\mathbf{\tilde e}_q$ be the representations of $v_p$ and $v_q$ from the two feed-forward networks with shared weights in the Siamese network. The probability between $v_p$ and $v_q$ is achieved via
\begin{equation}
    \mathcal{P}(v_p,v_q)=\sigma\left(W\left[\mathbf{\tilde{e}}_i:\mathbf{\tilde{e}}_j:\mathbf{\tilde{e}}_i-\mathbf{\tilde{e}}_j:\mathbf{\tilde{e}}_i\odot\mathbf{\tilde{e}}_j:1\right]\right)
\end{equation}
where $\sigma$ represents the sigmoid operator and $\odot$ is Hadamard product. Finally, we used the cross-entropy loss to compute the training loss for the deep framework.
\section{Experiments}

We used LectureBank \cite{li2019should}, University Courses \cite{liang2017recovering}, and ML of MOOCs \cite{pan2017prerequisite} to evaluate the performance and compare to the state-of-the-art methods, including binary classification models (SVM, LR, NB, RF) \cite{pan2017prerequisite}, RefD \cite{liang2015measuring}, GAE \cite{li2019should}, VGAE \cite{li2019should}, PREREQ \cite{roy2019inferring}, CPRL \cite{jia2021heterogeneous}, and Conlearn \cite{sun2022conlearn}. Besides, we employed precision, recall, and F1-score as evaluation metrics to measure the performance.

For BERT, we leveraged a combination of course lecture information and Wikipedia for concept description extraction in the MOOC and University Courses datasets. For the LectureBank dataset, we utilized the text information from the Wikipedia URLs of each concept in the dataset. When extracting vectors from texts, the max token size of BERT was set to 256 for all three datasets. 

For our method, the parameter $k$ of $k$-WL was set to 2. We used Adam as the optimizer with a learning rate of 0.00002 for all experiments. The batch size was set to 256 for the MOOC dataset and LectureBank dataset and 512 for the University Course dataset. The models were trained for 4000 epochs for all experiments until the loss stabilized. As for the baseline methods, we used default parameters as in their original implementations.

For all three datasets, we selected 80\% of the concept prerequisite pairs as the training set and 20\% of the concept prerequisite pairs as the test set. Negative samples were generated by randomly selecting unrelated phrase pairs from the vocabulary, along with the reverse pairs of the original positive samples. The results are recorded in Table \ref{mytab1}.

\renewcommand{\arraystretch}{1.5}
\begin{table}
\centering
\caption{Comparison between the results of the baseline model and our model} \label{mytab1}
\resizebox{\textwidth}{!}{
\begin{tabular}{|c|c|ccccccccccc|}
\hline
Datasets                                                & Metric    & NB    & SVM   & LR    & RF    & RefD  & GAE   & VGAE  & PREREQ & CPRL  & ConLearn & Ours  \\
\hline
\multirow{3}{*}{MOOCs}                                  & Precision & 0.577 & 0.668 & 0.748 & 0.375 & 0.784 & 0.293 & 0.266 & 0.448  & 0.800 & 0.895    & \textbf{\textbf{0.915}} \\
                                                        & Recall    & 0.623 & 0.577 & 0.270 & 0.669 & 0.188 & 0.733 & 0.647 & 0.592  & 0.642 & 0.850    & \textbf{0.860} \\
                                                        & F1-score  & 0.599 & 0.619 & 0.397 & 0.481 & 0.303 & 0.419 & 0.377 & 0.510  & 0.712 & 0.872    & \textbf{0.887} \\
\hline
\multirow{3}{*}{LectureBank}                            & Precision & 0.670 & 0.857 & 0.744 & 0.855 & 0.666 & 0.462 & 0.417 & 0.590  & \textbf{0.861} & 0.831    & 0.857 \\
                                                        & Recall    & 0.640 & 0.692 & 0.744 & 0.681 & 0.228 & 0.811 & 0.575 & 0.502  & 0.858 & 0.960    & \textbf{0.960} \\
                                                        & F1-score  & 0.655 & 0.766 & 0.744 & 0.758 & 0.339 & 0.589 & 0.484 & 0.543  & 0.860 & 0.891    & \textbf{0.906} \\
\hline
\multirow{3}{*}{University Courses}                     & Precision & 0.478 & 0.796 & 0.595 & 0.739 & \textbf{0.919} & 0.450 & 0.470 & 0.468  & 0.689 & 0.611    & 0.822 \\  
                                                        & Recall    & 0.649 & 0.635 & 0.546 & 0.480 & 0.415 & 0.886 & 0.694 & 0.916  & 0.760 & \textbf{0.966}   & 0.74  \\
                                                        & F1-score  & 0.550 & 0.707 & 0.569 & 0.582 & 0.572 & 0.597 & 0.560 & 0.597  & 0.723 & 0.749    & \textbf{0.778} \\  
\hline
\end{tabular}
}
\end{table}

From Table \ref{mytab1}, we can draw the following observations: the four binary classifiers, i.e., NB, SVM, LR, and RF, perform weak on the three datasets due to hand-crafted features, as well as RefD; GAE, VGAE, and PREREQ exploit the prerequisite relation information between KCs, resulting in improved performance; the CPRL method fails to delve into the textual information behind the concepts, leading to better performance; finally, both our algorithm and ConLearn extract prior information using the large-scale language model BERT and yield the best evaluations. Importantly, compared to ConLearn, our method uses 2-WL to integrate the structural information of the graph deeply, resulting in the best performance in terms of F1-score. All observations manifest that the introduced WL test into direct GNN is effective for CPRP.

\section{Conclusion}
This paper proposes a directed graph neural network based on the Weisfeiler-Leman algorithm to address the CPRP problem. Our method leverages BERT for KC text embeddings and redefines the $k$-tuple in the directed KC graph. Then, the 2-WL test is implemented to train a permutation-equivariant GNN. With the KC representation from GNN, the Siamese network computes the prediction probability of a KC link. Extensive experiments on three datasets demonstrate the superiority of the proposed method, achieving a more advanced performance than the state-of-the-art approaches of CPRP. Our future work will consider more evaluation results and topological information on the graph.

\bibliography{pmlr-sample}






\end{document}